\documentclass{sig-alternate-2013}

\usepackage{array}
\usepackage{color}
\usepackage{multirow}
\usepackage[labelfont=bf]{caption}
\usepackage[normalem]{ulem}
\usepackage{amssymb}
\usepackage{url}
\newcommand\etal{{\it et~al.\ }}

\newcommand\fig{Figure~}
\newcommand\tb{Table~}

\begin{document}

\conferenceinfo{ICMR'15,}{June 23 - 26, 2015, Shanghai, China}
\copyrightetc{\copyright~2015 ACM. ISBN \the\acmcopyr}
\crdata{978-1-4503-3274-3/15/06...\$15.00\\
DOI:~http://dx.doi.org/10.1145/2671188.2749406}

\title{Evaluating Two-Stream CNN for Video Classification}
\numberofauthors{1} 
%
\author{
%
%
\alignauthor
Hao Ye, Zuxuan Wu, Rui-Wei Zhao, Xi Wang, Yu-Gang Jiang\thanks{Corresponding author.}, Xiangyang Xue\\
\affaddr{School of Computer Science, Shanghai Key Lab of Intelligent Information 
Processing, \\ Fudan University, Shanghai, China}
\email{\{haoye10, zxwu,rwzhao14, xwang10, ygj, xyxue\}@fudan.edu.cn}
}

\maketitle
\begin{abstract}
Videos contain very rich semantic information. Traditional hand-crafted features are known to be inadequate in analyzing complex video semantics. Inspired by the huge success of the deep learning methods in analyzing image, audio and text data, significant  efforts are recently being devoted to the design of deep nets for video analytics. Among the many practical needs, classifying videos (or video clips) based on their major semantic categories (e.g., ``skiing") is useful in many applications. In this paper, we conduct an in-depth study to investigate important implementation options that may affect the performance of deep nets on video classification. Our evaluations are conducted on top of a recent two-stream convolutional neural network (CNN) pipeline, which uses both static frames and motion optical flows, and has demonstrated competitive performance against the state-of-the-art methods. In order to gain insights and to arrive at a practical guideline, many important options are studied, including network architectures, model fusion, learning parameters and the final prediction methods. Based on the evaluations, very competitive results are attained on two popular video classification benchmarks. We hope that the discussions and conclusions from this work can help researchers in related fields to quickly set up a good basis for further investigations along this very promising direction.
\end{abstract}




\section{Introduction}
\label{sec:introduction}


With the popularity of video recording devices and content sharing activities, there is a strong need for techniques that can automatically analyze the huge scale of video data. Video classification serves as a fundamental and essential step in the process of analyzing the video contents. For example, it would be extremely helpful if the massive consumer videos on the Web could be automatically classified into pre-defined categories. Learning semantics from the complicated video contents is never an easy task, and methods based on traditional hand-crafted features and prediction models are known to be inadequate~\cite{bengio2013representation}.

In recent years, deep learning based models have been proved to be more competitive than the traditional methods on solving complex learning problems in various domains. For example, the deep neural network~(DNN) has been successfully used for acoustic modeling in the large vocabulary speech recognition problems~\cite{Dahl:2012dx}. Moreover, the deep learning based methods have been shown to be extremely powerful in the image domain. In 2012, Krizhevsky \etal were the first to use a completely end-to-end deep convolutional neural network (CNN) model to win the famous ImageNet Large Scale Visual Recognition Challenge (ILSVRC), outperforming all the traditional methods by a large margin~\cite{krizhevsky2012imagenet}. In the most recent 2014 edition of the ILSVRC, Szegedy \etal developed an improved and deeper version of the CNN, which further reduced the top-5 label error rate to just 6.7\% over one thousand categories~\cite{Szegedy:2014tb}. In the text domain, deep models have also been successfully used for sentence parsing, sentiment prediction and language translation problems \cite{Socher:2011wx,Mikolov:2013uz,Kalchbrenner:2014wl,Sutskever:2014ty}.

On video data, however, deep learning often demonstrated worse results than the traditional techniques \cite{DBLP:conf/icml/JiXYY10,KarpathyCVPR14}. This is largely due to the difficulties in modeling the unique characteristics of the videos. On one hand, the spatial-temporal nature demands more complex network structures and maybe also advanced learning methods. One the other hand, so far there is very limited amount of training data with manual annotations in the video domain, which limits the progress of developing new methods as neural networks normally require extensive training. Very recently, Simonyan \etal proposed two-stream CNN, an effective approach that trains two CNNs using static frame and temporal motion separately~\cite{DBLP:conf/nips/SimonyanZ14}. The temporal motion stream is converted to successive optical flow images so that the conventional CNN designed for images can be directly deployed. 

Although promising results were observed in~\cite{DBLP:conf/nips/SimonyanZ14}, we underline that the performance of deep learning in video classification is subject to many implementation options, and there are no in-depth and systematic investigations on this in the field. In this paper, we conduct extensive experiments on two popular benchmarks to evaluate several important options, including not only network structures and learning parameters, but also model fusion that combines results from different networks and prediction strategies that map network outputs to classification labels. The two-stream CNN approach is adopted as the basic pipeline for the evaluations in this work. By evaluating the implementation options, we intend to answer the question of what network settings and implementation options are likely to produce good video classification results. As implementing a deep learning based system for video classification is a very difficult task, we hope that the discussions and conclusions from this work are helpful for researchers in this field and can stimulate future studies.

The rest of this paper is organized as follows. Section 2 reviews related works. In Section 3, we briefly introduce the two-stream CNN approach. Section 4 discusses the evaluated implementation options and Section 5 reports and analyzes experimental results. Finally, Section 6 summarizes the findings in this work.

\section{Related Works}
\label{sec:related}
Extensive studies have been conducted on video classification in the multimedia and computer vision communities. State-of-the-art video classification systems are usually built on top of multiple discriminative feature representations. To achieve better performance, various features have been developed. For instance, Laptev \etal extended the traditional SIFT features to obtain the Space-Time Interest Points (STIP) by finding representative tubes in 3D space~\cite{laptev2008learning}. Wang \etal proposed the dense trajectory features, which densely sample local patches from each frame at different scales and then track them in a dense optical flow field over time~\cite{wang2013action}. Besides the feature descriptors, one can obtain further improvements by adopting advantageous feature encoding strategies like the Fisher Vector~\cite{oneata2013action} or utilizing fusion techniques~\cite{snoek2005early,ye2012robust,mm14:videoclassification} to integrate information from different features. 

The aforementioned hand-crafted features like the dense trajectories have demonstrated state-of-the-art performance on many video classification tasks. However, these features are still unsatisfying and the room for further improvements may be limited. In contrast to the hand-crafted features, there is a growing trend of learning features directly from raw data using deep learning methods, among which the CNN~\cite{lecun-01a} has attracted wide attentions due to their great success in image classification~\cite{krizhevsky2012imagenet,Szegedy:2014tb}, visual object detection~\cite{girshick2014rcnn}, etc.

Compared with the extensive studies on using deep learning for image analysis, only a few works have exploited this approach for video analysis. Ji \etal and Karparthy \etal extended the CNN into temporal domain by stacking static frames, upon which convolution can be performed with space-temporal filters~\cite{DBLP:conf/icml/JiXYY10,KarpathyCVPR14}. However, the learned representations from these methods produced worse results than the state-of-the-art hand-crafted features like the dense trajectories~\cite{wang2013action}. More recently, Simonyan \etal \cite{DBLP:conf/nips/SimonyanZ14} achieved very competitive performance by training two CNNs on spatial (static frames) and temporal (optical flows) streams separately and then fusing the two networks. 

Most CNN-based approaches rely on the neural networks to perform the final class label prediction, normally using a softmax layer~\cite{KarpathyCVPR14,DBLP:conf/nips/SimonyanZ14} or a linear layer~\cite{DBLP:conf/icml/JiXYY10}. Instead of direct prediction by the network, Jain \etal conducted action recognition using support vector machines (SVMs) with features extracted from off-the-shelf CNN models~\cite{JainTHUMOS14}. Their impressive results in the THUMOS action recognition challenge~\cite{THUMOS14} indicate that CNN features are very powerful. In addition, a few works attempted to apply the CNN representations with Recurrent Neural Network (RNN) models to capture the temporal information in videos and perform classification within the same network. Donahue \etal leveraged the Long-Short Term Memory (LSTM) RNN model for action recognition~\cite{DBLP:journals/corr/DonahueHGRVSD14} and Venugopalan \etal proposed to translate videos directly to sentences with the LSTM model by transferring knowledge from image description tasks~\cite{DBLP:journals/corr/VenugopalanXDRMS14}. RNN shares the same motivation with the temporal pathway in the two-stream framework~\cite{DBLP:conf/nips/SimonyanZ14}.   

In this paper, we provide an in-depth study on various implementation choices of deep learning based video classification. On top of the two-stream CNN pipeline~\cite{DBLP:conf/nips/SimonyanZ14}, we examine the performance of the spatial and temporal streams separately and jointly with different network architectures under various sets of parameters. In addition, we also examine the effect of different prediction options, including directly using a CNN with a softmax layer for end-to-end classification and adopting SVMs on the features learned from the CNNs. The RNN models are not considered since they have not been fully explored and only demonstrated limited performance gain in the context of video classification.  This paper fills the gap in the existing works on deep learning for video classification, where most people have focused on designing a new classification pipeline or a deeper network structure without systematically evaluating and discussing the implementation options.  

\begin{figure*}[ht]
\centering
\includegraphics[scale=0.9]{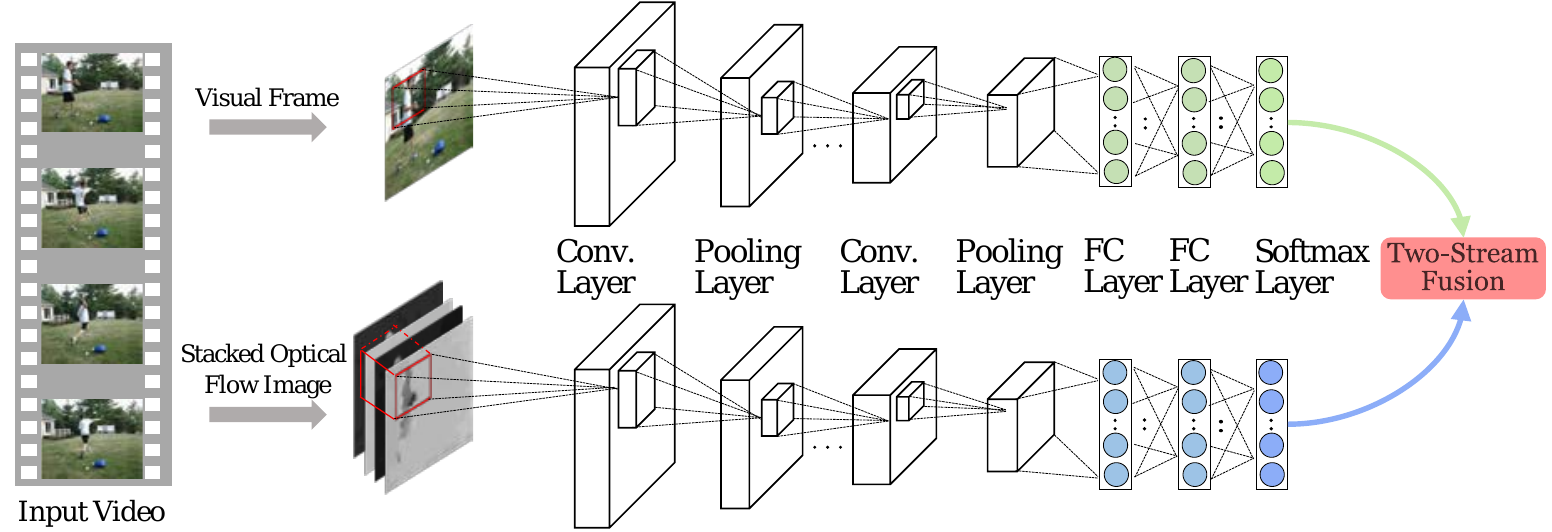}
\vspace{-0.06in}
\caption{The processing pipeline of the two-stream CNN \cite{DBLP:conf/nips/SimonyanZ14}, which has demonstrated competitive video classification performance. ``Conv." indicates  ``Convolution" and ``FC" is the abbreviation of ``Fully Connected". Our evaluations in this work are conducted on top of this pipeline.}
\label{fig:twostream}
\vspace{-0.08in}
\end{figure*}

\section{Two-Stream CNN}
\label{sec:twostream}
According to the findings in neuroscience, human visual system processes what we see through two kinds of streams, namely the ventral pathway and the dorsal pathway respectively. The ventral pathway is responsible for processing the spatial information, such as shape and color, while the dorsal pathway is responsible for processing the motion information~\cite{Kruger:2013gc,Goodale:1992wc}. Mimicking the human vision mechanism, the video data, i.e., sequential image frames, can be naturally decomposed into the spatial and temporal components in a similar way. More specifically, image content including scenes and objects belong to the spatial component. The complementary temporal component contains motion information across video frames. As an early attempt, Schindler \etal extracted spatial and temporal features on still images and optical flows by using the Gabor filters~\cite{schindler2008action}. With the recent success of the CNN, as briefly mentioned earlier, Simonyan \etal promoted this framework by the two-stream CNN structure that models both the spatial and the temporal information~\cite{DBLP:conf/nips/SimonyanZ14}. \fig \ref{fig:twostream} shows the processing pipeline of this approach. One CNN is applied to process the spatial stream of the video data, and the other CNN is used to handle the temporal stream.

The processing data flow of the spatial stream is shown on top of \fig \ref{fig:twostream}. This CNN has the same structure as a general deep CNN for image classification~\cite{Zeiler:2011hy,krizhevsky2012imagenet,Szegedy:2014tb}. It directly takes individual video frames as network inputs followed by several convolutional layers, pooling layers and fully connected (FC) layers. Finally, after a softmax layer, the network outputs in the range of [0, 1] are taken as the predicted probabilities of the video classes. Multiple frames are sampled from each input video. The network processes one frame each time and the predictions on individual frames are merged simply by average probability fusion.

For the temporal stream, in order to capture the motion information, the temporal CNN takes stacked optical flows as input, rather than the frames in the spatial stream. Specifically, dense optical flow can be seen as a set of displacement vector fields between consecutive frames. For a given pair of consecutive frames, the horizontal and vertical components of the calculated displacement vector fields between them can be used to generate two ``optical flow images", respectively. To further consider temporal information, one can stack the optical flow images of each frame at time $t$ and its $L$ subsequent frames into a stacked $2L$-channel optical flow image (in contrast to the traditional 3-channel RGB images). The network architecture and training process of the temporal CNN are basically the same as the spatial counterpart, except that the input images have a different number of channels. There are multiple stacked $2L$-channel optical flow images in a video. The way of fusing predictions on these individual images is also the same as that of the spatial channel.

We now have predictions from the two CNNs, based on the spatial and temporal streams separately.  The last step is to combine the two streams to produce the final output. For this, linear fusion with fixed weights was used in~\cite{DBLP:conf/nips/SimonyanZ14}.

\section{Implementation Options}
In this section, we discuss several important implementation options that can affect the performance of deep learning  for video classification, including network architectures, fusion strategies, learning parameters and prediction options. 

\subsection{Network Architectures}
Network architecture plays a very important role in the performance of a deep learning model. The current surge of the CNNs in many tasks heavily relies on the use of superior network architectures, such as the AlexNet~\cite{krizhevsky2012imagenet} and the GoogLeNet~\cite{Szegedy:2014tb}. Popular CNN architectures usually have alternating convolutional layers and auxiliary layers (e.g., pooling layers), and are toped by a few fully connected (FC) layers. Recent studies demonstrate that better recognition accuracies can be achieved by deepening the CNN architectures~\cite{simonyan2014very,Szegedy:2014tb}, which means that deeper architectures can lead to progressively more discriminative features at higher layers. To evaluate the performance of different network architectures, in this work, we adopt and evaluate a \emph{medium} network structure, CNN\_M~\cite{DBLP:conf/nips/SimonyanZ14}, and a very recent \emph{deeper} network architecture, VGG\_19~\cite{simonyan2014very}. See \tb\ref{tb:archi} for the detailed configurations of CNN\_M and VGG\_19.

\paragraph{CNN\_M Network}
CNN\_M basically follows the same spirit as the widely adopted AlexNet~\cite{krizhevsky2012imagenet}. It contains five convolutional layers followed by three FC layers and the input image is fixed to the size of $224 \times 224$. Compared with AlexNet~\cite{krizhevsky2012imagenet}, CNN\_M possesses more convolutional filters. On the first convolutional layer, the size and stride are both smaller ($7 \times 7$ and 2 respectively) than those in AlexNet, while the remaining convolutional layers have the same filter size and stride. By increasing the number of filters and reducing the filter size and the stride step, CNN\_M can discover more subtle information from input images, and hence can obtain more robust feature representations and better predictions. Note that CNN\_M offers a $13.5$\% top-5 error on the ILSVRC-2012 validation set~\cite{DBLP:conf/nips/SimonyanZ14} (a famous image recognition benchmark), which is generally considered as a good result.

\paragraph{VGG\_19 Network}
VGG\_19 not only further reduces the size of convolutional filters and the stride, but more importantly, it also extends the depth of the network. More precisely, VGG\_19 consists of nineteen layers, including sixteen convolutional layers and three fully connected layers. In addition, the size of all the convolutional filters decreases to $3 \times 3$ and the stride reduces to only 1 pixel, which enables the network to explore finer-grained details from the feature maps. With this much deeper architecture, VGG\_19 possess strong capabilities of learning more discriminative features and the high-level final predictions. It can produce a $7.1$\% top-5 error rate on the ILSVRC-2012 validation set~\cite{simonyan2014very}.

\newcommand{\sr}{\rule[-0.15cm]{0pt}{0.6cm}}
\newcolumntype{x}[1]{>{\centering\arraybackslash\hspace{0pt}}p{#1}}
\begin{table}[t!]
\centering
\begin{tabular}{|c|x{3.6cm}|x{3cm}|}
\hline 
  & CNN\_M & VGG\_19 \\
\hline 
\hline 
\multirow{21}{*}{\rotatebox{90}{Convolutional Layers}}
										   & 	96  $\times$ 7 $\times$ 7    &  64  $\times$ 3 $\times$ 3 \\ 
										   & 	(stride: 2, padding: 0)  	     &  64  $\times$ 3 $\times$ 3 \\ 
										   & 	LRN 				 	     &     	$\times 2$  pooling    \\ 
										   & 	   $\times 2$  pooling       &  128 $\times$ 3 $\times$ 3 \\ 
										   & 	256 $\times$ 5 $\times$ 5    &  128 $\times$ 3 $\times$ 3 \\ 
										   & 	(stride: 2, padding: 1) 	     &      $\times 2$ pooling    \\ 
										   & 	LRN  					     &  256 $\times$ 3 $\times$ 3 \\ 
										   & 	   $\times 2$ pooling  	 	 &  256 $\times$ 3 $\times$ 3 \\ 
										   & 	512 $\times$ 3 $\times$ 3    &  256 $\times$ 3 $\times$ 3 \\ 
										   & 	(stride: 1, padding: 1) 	     &  256 $\times$ 3 $\times$ 3 \\ 
										   & 	512 $\times$ 3 $\times$ 3    &      $\times 2$  pooling    \\ 
										   & 	(stride: 1, padding: 1) 	     &  512 $\times$ 3 $\times$ 3 \\ 
										   & 	512 $\times$ 3 $\times$ 3     &  512 $\times$ 3 $\times$ 3 \\ 
										   & 	(stride: 1, padding: 1)		 	     &  512 $\times$ 3 $\times$ 3 \\ 
										   & 	$\times 2$ pooling 	     &  512 $\times$ 3 $\times$ 3 \\ 
										   & 				 	     &      $\times 2$  pooling    \\ 
										   & 						 	     &  512 $\times$ 3 $\times$ 3 \\ 
										   & 						 	     &  512 $\times$ 3 $\times$ 3 \\ 
										   & 						 	     &  512 $\times$ 3 $\times$ 3 \\ 
										   & 						 	     &  512 $\times$ 3 $\times$ 3 \\ 
										   & 						 	     &      $\times 2$  pooling    \\ \hline
\multirow{3}[5]{*}{\rotatebox{90}{FC Layers}} 
										   & \sr 4,096 neurons  & \sr 4,096 neurons \\ \cline{2-3}
									       & \sr 2,048 neurons  & \sr 4,096 neurons \\ \cline{2-3}
									       & \sr softmax  & \sr softmax \\ \hline
\end{tabular}
\caption{ Configurations of two networks: CNN\_M and VGG\_19. The convolutional kernel parameter is denoted in the form ``\# filters $\times$ kernel size $x$ $\times$ kernel size $y$''. For both networks, max-pooling with a sampling factor of 2 is used (``$\times 2$ pooling"). In VGG\_19, both stride and padding are set to 1 for all the convolutional layers. The VGG\_19 does not contain the Local Response Normalization (LRN), which requires considerable computation but contributes little to the performance. }
\label{tb:archi}
\end{table}

\subsection{Fusion Strategies}
Fusing multiple clues is a standard technique in video analysis, which can often lead to better performance. We divide this part into model fusion and modality fusion.

\subsubsection{Model Fusion}
Combining various deep learning models can usually produce better performance than using just a single model, because models using different architectures or trained with different parameters may contain complementary information. For instance, a model trained with larger convolutional filters may focus more on large patterns, while one with smaller filters may be more sensitive to finer-grained details. Since both the two CNN architectures, CNN\_M and VGG\_19, can be trained with multiple parameters, there are consequently several candidate models that can be exploited in this experiment. We examine different combinations to integrate information from these candidate models. 

\subsubsection{Modality Fusion}
As aforementioned, videos are naturally multimodal, and hence the integration of the spatial and the temporal streams is very important. Simonyan \etal \cite{DBLP:conf/nips/SimonyanZ14} adopted a simple linear fusion method and fixed weights without explanations. We will examine the effect of this fusion weight on two very different datasets.  Notice that another very important modality, the audio channel, is not considered in this work because we focus on examining options of training models using only the visual data following the two-stream pipeline. However, we strongly believe that better performance can be achieved by further including the audio information, because this has been observed in many recent works~\cite{icmr11:consumervideo}.

\subsection{Learning Parameters}
Although deep learning has achieved promising results on many tasks, training and fine-tuning a good model normally require significant efforts as there are several important parameters that need to be evaluated, such as learning rate, dropout ratio and the number of training iterations. These seemingly arbitrarily chosen parameters can influence the performance significantly. For instance, a small learning rate may demand much more iterations to converge, while a large value may accelerate the convergence but can possibly result in oscillation. In addition, a larger dropout ratio may lead to a better model but could probably slow down the convergence. There is no universal rule for parameter selection. With the goal of providing some insights on  parameter selection specially for the problem of video classification, we study different sets of parameters using the two aforementioned network architectures and two datasets. 

In particular, the dropout ratio and the number of iterations are jointly evaluated and discussed. For learning rate, we set it to $10^{-2}$ initially, and then decreased to $10^{-3}$ after 100K iterations, then to $10^{-4}$ after 200K iterations. In the fine-tuning case, the rate starts from $10^{-3}$ and decreases to $10^{-4}$ after 14K iterations, then to $10^{-5}$ after 20K iterations. This setting is similar to~\cite{DBLP:conf/nips/SimonyanZ14}, but we start from a smaller rate of $10^{-3}$ instead of $10^{-2}$. Note that other choices on learning rate are not evaluated as we find that the final performance is less sensitive to this parameter as long as it is set following the suggested rules in~\cite{DBLP:journals/corr/Schmidhuber14}.

\subsection{Prediction Options} 
While neural networks can act as end-to-end classifiers by using a final softmax layer, traditional classifiers like the SVMs can also be deployed on the features extracted by the CNN, which are generally the outputs of the last several fully connected layers. Recently, Razavian \emph{et al.}~\cite{DBLP:journals/corr/RazavianASC14} adopted features extracted from a CNN model pre-trained on ImageNet to perform classification with SVMs. They demonstrated strong performance on image analysis tasks like scene recognition, object detection, etc. In addition, Jain \etal \cite{JainTHUMOS14} leveraged the CNN features using SVMs for action recognition in videos, and achieved superior performance on the THUMOS action recognition benchmark~\cite{THUMOS14}. The results suggest that the CNN features may be used in combination with traditional classifiers for improved performance, but these existing works were performed only on images or the spatial frames. This paper investigates the performance of features extracted from different layers of both the spatial and the temporal CNNs using SVMs for classification. Results are compared with that of the end-to-end neural network based approach. 

\section{Experiments}
\subsection{Datasets and Evaluation Criteria}
\textbf{UCF-101}~\cite{ucf101}. The UCF-101 dataset is a widely adopted benchmark for action recognition in videos, which consists of 13,320 video clips (27 hours in total). There are 101 annotated classes that can be divided into five types: Human-Object Interaction, Body-Motion Only, Human-Human Interaction, Playing Musical Instruments and Sports. We perform evaluation according to the popular 3 train/test splits following~\cite{THUMOS14}. Results are measured by classification accuracy on each split and mean accuracy over the 3 splits. For some evaluations, we only report results on the first split due to computation limitation.

\textbf{Columbia Consumer Videos (CCV)}~\cite{icmr11:consumervideo}. The CCV dataset contains 9,317 YouTube videos annotated according to 20 classes, such as ``wedding ceremony'', ``birthday party'', ``skiing'' and ``playground''. We follow the protocol defined in~\cite{icmr11:consumervideo} to use a training set of 4,659 videos and a test set of 4,658 videos. Results are measured by average precision (AP) for each class and mean AP (MAP) across all the classes~\cite{icmr11:consumervideo}.
Note that, different from the UCF101 actions, most classes in CCV are social events, sports, objects and scenes. In addition, the average duration of this dataset is around 80 seconds, which is over ten times longer than that of the UCF-101. We hope that using the two datasets with different characteristics can help lead to more generalizable conclusions.

For both datasets, we adopt the same data augmentation strategies as \cite{DBLP:conf/nips/SimonyanZ14}.

\subsection{Network Options}
Using what network structure is the first decision we have to make in the implementation of a deep learning based video classification system. There are numerous options. In this work, we evaluate and compare the two popular structures CNN\_M and VGG\_19. 

Results of the spatial stream are summarized in \tb \ref{tb:networkstructure}.  As can be seen, VGG\_19 produces consistently better results on both datasets, indicating that larger (deeper) networks are generally better. This is consistent with the observations on the large scale image classification tasks. Results of different dropout rates are listed in this table because this can ofter a more complete understanding of the power of the networks under different  settings of learning parameters. Detailed discussions on the effect of dropout rates will be given later. 

For the temporal stream, we also tried to use the VGG\_19 network under a few parameter settings, but observed clearly worse results than the CNN\_M. As the gap is clear, we did not proceed to finish all the parameters to fully compare the two networks. The key reason is that the temporal stream has to be trained from scratch, which is different from the spatial stream that can use fine-tuning to adjust the pre-trained network based on millions of images \cite{DBLP:conf/nips/SimonyanZ14}. In this case, learning a smaller temporal network is more feasible with limited training data. We expect that better results can be achieved by VGG\_19 for the temporal stream when there is sufficient training data.

\begin{table}[t!]
\centering
\begin{tabular}{|c|c|c|}
\hline
Models      & UCF-101 (split-1)    & CCV     \\ \hline \hline
CNN\_M dr1  & 71.58\% & 68.78\% \\ \hline
CNN\_M dr3  & 68.65\% & 68.81\% \\ \hline
CNN\_M dr5  & 68.25\% & 68.64\% \\ \hline
CNN\_M dr7  & 68.15\% & 67.40\% \\ \hline
CNN\_M dr9  & 60.85\% & 51.81\% \\ \hline \hline
VGG\_19 dr1 & 75.87\% & 74.66\% \\ \hline
VGG\_19 dr3 & 79.59\% & 74.47\% \\ \hline
VGG\_19 dr5 & 80.41\% & 75.04\% \\ \hline
VGG\_19 dr7 & 76.66\% & 74.90\% \\ \hline
VGG\_19 dr9 & 76.39\% & 73.09\% \\ \hline
\end{tabular}
\caption{Spatial stream results of two network architectures on UCF-101 and CCV under different dropout rates (``dr1" indicates the 0.1 dropout rate). See texts for discussions.}
\label{tb:networkstructure}
\end{table}

\subsection{Model and Modality Fusion}
Next, we discuss results by fusing models and modalities. For the combination of different models, we use the spatial stream and 10 network models (2 structures each trained with 5 dropout rates). We tried all the possible modal combinations by averaging their prediction scores and identified the top 3 results in order to learn which model is more reliable and what combinations are good. Results are shown in \tb \ref{tb:modalfusion}. We see that VGG\_19 models are more ``popular" in the top combinations on both datasets, confirming the fact that fusing good models generally offers good results. However, comparing the model fusion results with the individual model results in \tb \ref{tb:networkstructure}, it is clear that fusing models does not improve results significantly. For instance, on the UCF-101, the 2nd best results by fusing five models is actually the same with just using the best single model (VGG\_19 dr5). Therefore, one conclusion is that fusing a strong network (VGG\_19) with a relatively weaker network (CNN\_M) does not help much (if not becoming worse). Notice that average model prediction fusion is adopted here. Using dynamic fusion weights may lead to better results, but the gain is unlikely to be significant. 

\begin{table}[t!]
\small
\centering
\begin{tabular}{|c|c|c|c|c|c|c|}
\hline
\multirow{2}{*}{Models}
  					  &  \multicolumn{3}{c|}{UCF-101 (split-1) }           & \multicolumn{3}{c|}{CCV}      	     \\ \cline{2-7}
   					  &  1       		 & 2		  &  3   		     &  1         & 2           & 3     	\\ \hline \hline
CNN\_M   dr1          &  \checkmark      & \checkmark &                  &  		  &			    & \checkmark \\ \hline
CNN\_M   dr3          &  			     &            &                  &  		  &			    &			 \\ \hline
CNN\_M   dr5          &  		         &            &                  &  		  &			    &			 \\ \hline
CNN\_M   dr7          &                  &            &                  &            &			    &			\\ \hline
CNN\_M   dr9          &                  & \checkmark &  \checkmark      & \checkmark &			   &			 \\ \hline
VGG\_19  dr1          &  			     &            &                  &  		  &			    & \checkmark \\ \hline
VGG\_19  dr3          &  \checkmark      & \checkmark &                  &  		  &  \checkmark & \checkmark \\ \hline
VGG\_19  dr5          &  \checkmark      & \checkmark &  \checkmark      & \checkmark &  \checkmark & 			 \\ \hline
VGG\_19  dr7          &  			     & 			  &  		         & \checkmark &  \checkmark & \checkmark \\ \hline
VGG\_19  dr9          &  \checkmark	     & \checkmark &  		         &  		  &  		    & \checkmark \\ \hline \hline
Perf. (\%)      		  &  80.46           & 80.41      & 80.33            & 75.31 	  &  75.31      & 75.30      \\ \hline
\end{tabular}
\caption{Top-3 spatial stream model fusion results on both datasets. The ``\checkmark" sign indicates the used models in each of the top combinations. A general observation is that fusing good models like the VGG\_19 based ones tend to generate good results, but the improvement is fairly limited.}
\label{tb:modalfusion}
\end{table}

\begin{figure}
\centering \includegraphics[scale=0.48]{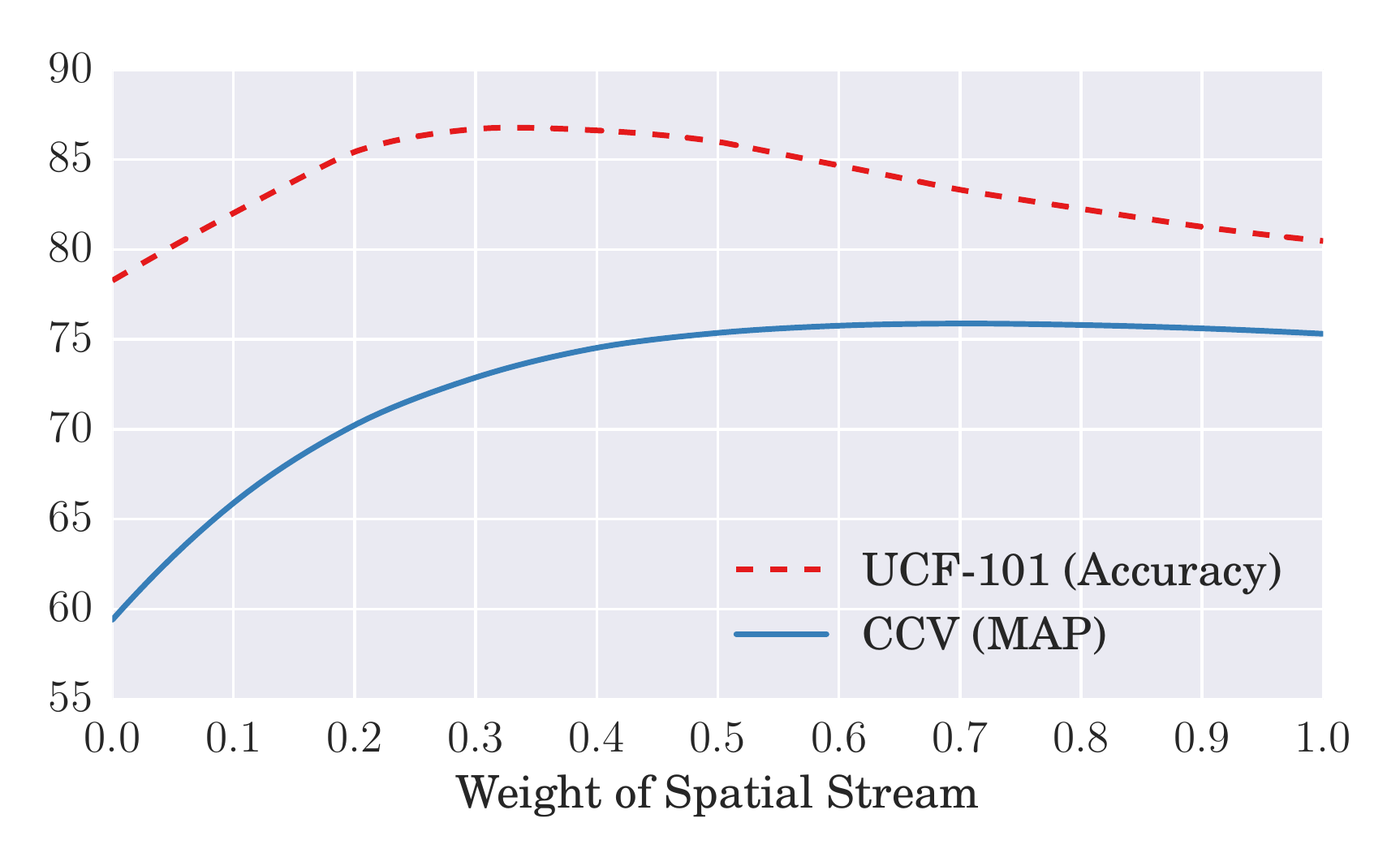}
\vspace{-0.15in}
\caption{Performance (\%) of combining spatial and temporal streams using linear fusion with different weights. Temporal stream weight is set as ``$1-$spatial weight". Thus the left-end points of the curves indicate the performance of using the temporal stream alone, while the right-end is the spatial stream performance.}
\label{fg:modalityfusion}
\end{figure}

\begin{table*}[t!]
\centering
\begin{tabular}{|c|c|c|c|c|c||c|c|}
\hline
				& 			  &\multicolumn{2}{c|}{Spatial (VGG\_19)}      & \multicolumn{2}{c||}{Temporal (CNN\_M)} 				  &\multicolumn{2}{c|}{Spatial-Temporal Fusion} \\ \cline{2-8}
                & Feature     & Early Fusion	    & Late Fusion     & Early Fusion    & Late Fusion &  Early Fusion  & Late Fusion \\ \hline \hline

\multirow{3}{*}{\hfil UCF-101 (split-1)} 
			    & FC1      	  &  75.84\%     & 70.71\%    &  78.22\%   & 76.74\%  & 87.55\%  & 82.34\%    \\ \cline{2-8} 
                & FC2         &  72.75\%     & 64.42\%    &  77.85\%   & 76.24\%  & 85.94\%  & 80.52\%    \\ \cline{2-8} 
                & FC1\&FC2  &  75.73\%     & 70.29\%    &  78.30\%   & 76.63\%  & 87.71\%  & 82.00\%    \\ \hline \hline
                                                                                                                                                                  
\multirow{3}{*}{\hfil CCV}  
				        & FC1        &  70.75\%     & 67.34\%    &  58.04\%   & 54.08\%  & 73.25\%  & 68.87\%    \\ \cline{2-8} 
                & FC2        &  70.45\%     & 68.85\%    &  55.86\%   & 52.52\%  & 72.76\%  & 70.06\%    \\ \cline{2-8} 
                & FC1\&FC2   &  71.15\%     & 69.25\%    &  58.79\%   & 54.40\%  & 73.27\%  & 69.90\%    \\ \hline\end{tabular}
\caption{Prediction results of SVMs classifiers on the CNN features. FC1 (FC2) indicates the output feature of the first (second) fully connected layer. ``FC1\&FC2" is the concatenation of the FC1 and FC2 features. }
\label{tb:SVM} 
\end{table*}

For modality fusion, we use the best spatial network configurations to fuse with a temporal network trained using the CNN\_M architecture. Results of different fusion weights are plotted in \fig \ref{fg:modalityfusion}. Comparing the temporal stream with the spatial counterpart, the latter produces better results on both datasets. The gap is larger on CCV as its categories are easier to be recognized by viewing just one or a few frames, e.g., the sports classes ``basketball" and ``skiing". Fusing the two modalities is effective, leading to significant improvement on UCF-101 (best fusion: 86.7\%; spatial: 80.5\%; temporal: 78.3\%). The gain on CCV is limited as the result of the temporal stream is not good (best fusion: 75.9\%; spatial: 75.3\%; temporal: 59.4\%), which is generally consistent with the results of the hand-crafted features on this dataset~\cite{icmr11:consumervideo}\footnote{On CCV, it was reported that static frame features are significantly better than spatial-temporal features.}. As for the suitable fusion weight, the results indicate that similar or higher temporal weight is preferred for classes that can be better recognized by viewing a clip (not just a frame), even when the temporal stream performs worse than the spatial stream.

\subsection{Effect of Learning Parameters}
\begin{figure*}[t!]
\centering \includegraphics[scale=0.48]{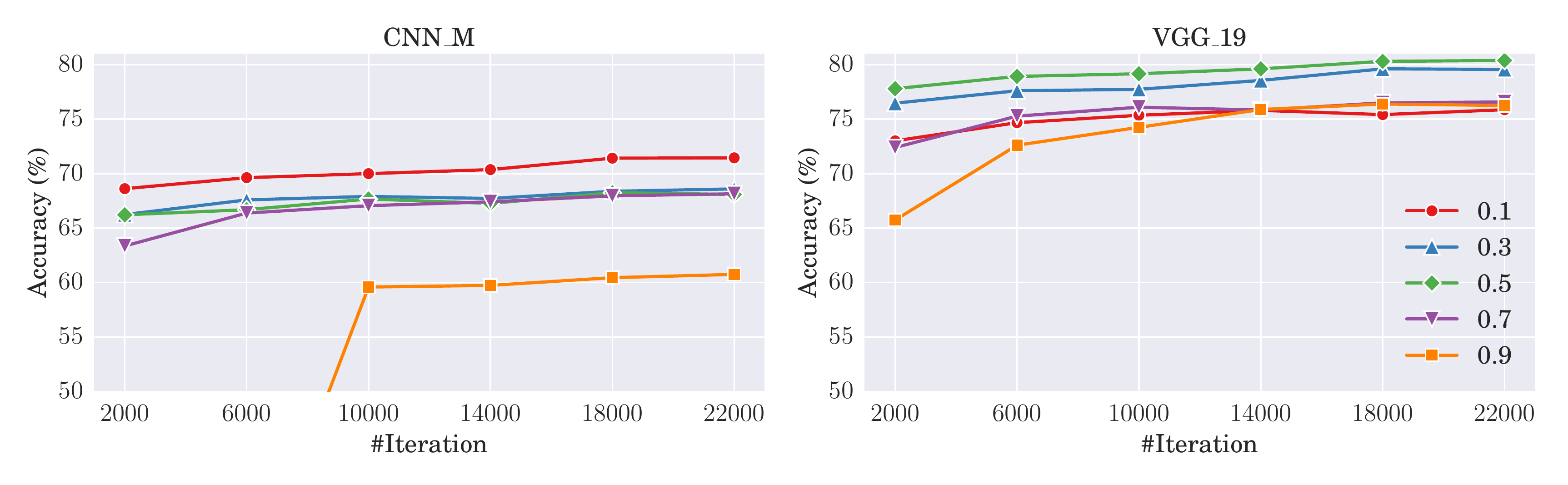}
\caption{Spatial stream results on UCF-101 (split-1), under different dropout ratios (from 0.1 to 0.9) and iteration numbers.}
\label{fg:spatialparaUCF}
\end{figure*}

\begin{figure*}[t!]
\centering \includegraphics[scale=0.48]{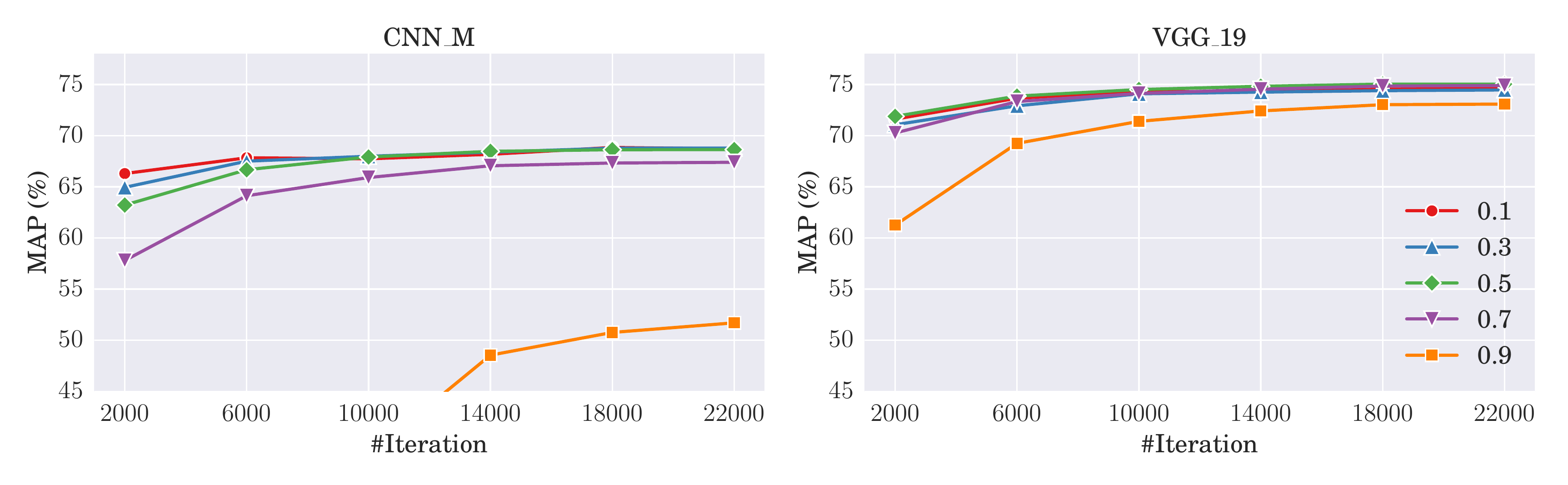}
\caption{Spatial stream results on CCV, under different dropout ratios and iteration numbers.}
\label{fg:spatialparaCCV}
\end{figure*}

We jointly evaluate the effect of two learning parameters: the dropout ratio and the number of training iterations.  We first study the spatial stream. \fig \ref{fg:spatialparaUCF} and \fig \ref{fg:spatialparaCCV} visualize the results on UCF-101 and CCV respectively. Notice that the spatial networks are fine-tuned based on the models pre-trained on the Image-Net (only the last three FC layers are fine-tuned), and therefore they start from a fairly good initialization and become stable quickly after just 10-20 thousands of iterations. With the uniform settings on learning rate (cf. Section 4.3), the number of iterations required to reach convergence is similar across different network architectures and dropout ratios. One interesting observation is that large dropout ratios (e.g., 0.9) are especially unsuited for small networks like the CNN\_M.  This is probably because a small network can hardly learn anything if as high as 90\% of the information are dropped at each iteration, particularly for the case of fine-tuning where only the last three layers are adjusted. 

\begin{figure*}[t!]
\centering \includegraphics[scale=0.48]{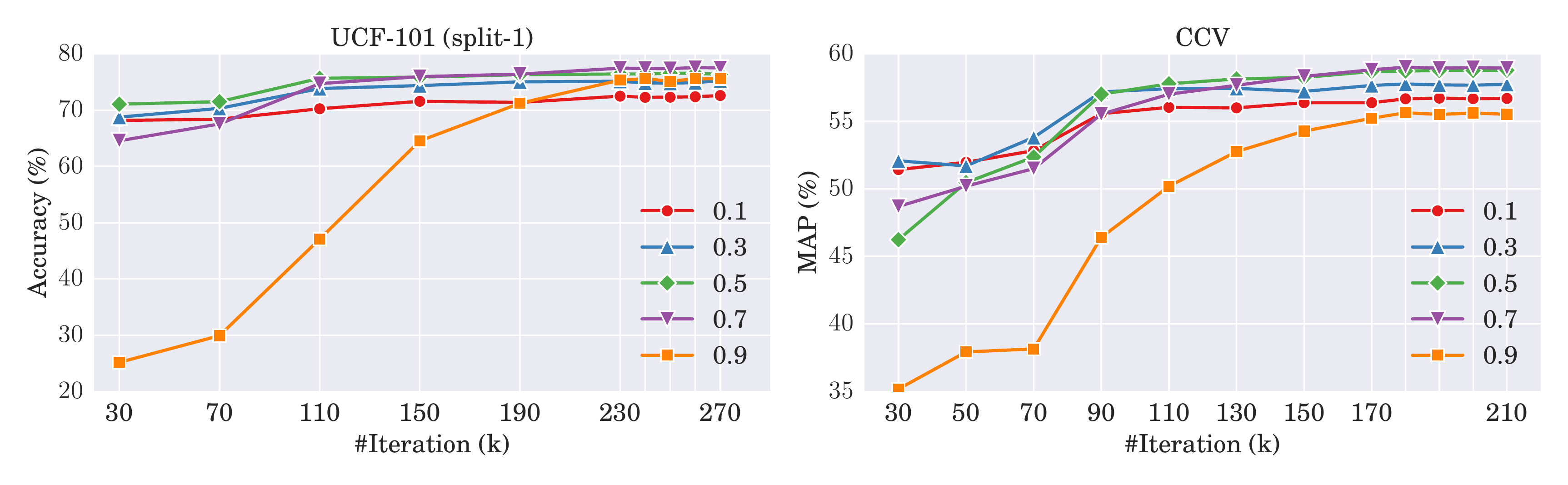}
\vspace{-0.1in}
\caption{Temporal stream results on both UCF-101 (split-1) and CCV, using CNN\_M with different dropout ratios and iteration numbers.}
\label{fg:temporalpara}
\end{figure*}

\fig \ref{fg:temporalpara} plots the temporal stream results on both datasets, using the CNN\_M architecture with various learning parameters. We observe that a large dropout ratio requires more iterations to reach a high level of performance, which is easy to understand. Different from the spatial stream observations on CNN\_M, a large dropout ratio can also lead to comparable results. This may be because the temporal stream networks are trained from scratch, and, even using the same architecture, training the entire framework is more complex than fine-tuning (only tuning three FC layers). Furthermore, some researchers expressed the view that dropout can be considered as a form of training set augmentation~\cite{DBLP:journals/corr/Schmidhuber14}. Complex networks may be more suitable to learn from highly augmented input data. 

\subsection{Softmax vs. SVMs}
As discussed in Section 4.3, neural networks can be used as an end-to-end classifier or a feature extractor. In this section, we discuss results of using SVMs on the CNN features, and compare with softmax. We train linear SVMs using outputs from the first and the second FC layers. VGG\_19 and CNN\_M are adopted for the spatial and temporal stream respectively. As each video has multiple spatial frames and stacked optical flow images, there are two ways to train the SVMs classifiers. One is early fusion by averaging features from all the frames first before classifier training and testing. The other is late fusion, which takes all the frames as inputs separately and uses average prediction scores as the final video-level score.

Table \ref{tb:SVM} summarizes the results of spatial, temporal, and their fusion on both datasets. Compared with the softmax based prediction, SVMs is only slightly better on the UCF-101 under the spatial-temporal modality fusion setting using the early  frame fusion method. On CCV, the performance is significantly lower than softmax. 
Comparing early frame fusion with late fusion, early fusion is consistently good, indicating that averaging frame features before classification may help suppress noises that affect SVMs training. Interestingly, the neural networks take individual frames as inputs like the late fusion based SVMs, but are quite robust.

\begin{table}[h!]
\centering
\begin{tabular}{|c|c||c|c|}
\hline
	\multicolumn{2}{|c||}{UCF-101}       & 	\multicolumn{2}{c|}{CCV}		\\ 	\hline \hline
Simonyan \etal\cite{DBLP:conf/nips/SimonyanZ14} &  88.0\%  & Ye \etal\cite{ye2012robust} & 64.0\% \\ \hline
Peng \etal\cite{peng2014bag}			&   87.9\%	 & Jhuo \etal\cite{MVA:audiovisual} & 64.0\% \\ \hline	    
Wang \etal\cite{wang2013action} 		& 	85.9\%	 & Liu \etal\cite{liu2013sample} & 68.2\%  \\ \hline		  	
Karpathy \etal\cite{KarpathyCVPR14} &65.4\%& Wu \etal\cite{mm14:videoclassification} &  69.3\% \\ \hline \hline    
Ours - Softmax                      			&   86.7\%  & Ours - Softmax  &  75.9\%		        \\  \hline
Ours - SVM                      			&   87.7\%  & Ours - SVM  &  73.3\%		        \\  \hline
\end{tabular}
\caption{Comparison with the state-of-the-art results.}
\label{tb:comparison} 
\end{table}

\subsection{Comparison with the State of the Arts}
Finally, we compare our results with the state of the arts on both datasets. For UCF-101, we report average accuracies over the three official train-test splits. As shown in \tb \ref{tb:comparison}, our results are competitive on both datasets. The performance on CCV is significantly better than the state of the arts, all of which adopted traditional features like the dense trajectories \cite{wang2013action}. On UCF-101, our results are comparable to a very recent work \cite{peng2014bag}, which uses an extensive fusion approach on top of state-of-the-art hand-crafted features. Our results are also similar to that of \cite{DBLP:conf/nips/SimonyanZ14}. Note that the 88.0\% reported in \cite{DBLP:conf/nips/SimonyanZ14} was attained by using \emph{external training data} from another human action benchmark. If only trained on UCF-101, the performance is lower.

\section{Summary and Discussion}
\label{set:summary}
Building a deep learning system for video classification is not an easy task. We have evaluated several important implementation options. The major findings are summarized in the following. Note that knowing what works and what does not work is equally important. 

For network architectures, one observation is that deeper networks like the VGG\_19 are better, but a sufficient amount of training data is required. This is fine for the spatial stream, as the image annotations like the ImageNet can be used to pre-train the network. For the temporal stream, as we cannot use the image collections for model pre-training, the results of very deep networks are not good. We envision that the they would work well on the temporal stream once we have a large amount of training data in the video domain.

Results indicate that fusing multiple network models is not very helpful, especially when combining a strong network with a weak one. Fusing two networks with a similar performance level but different architectures might help, but this is not verified based on our experiments. In addition, combining predictions from the spatial and the temporal streams is useful. This is important for the classification of long-term procedural actions, which benefits significantly from the temporal clues. The linear weighted spatial-temporal fusion method works well but is not ideal. This aspect deserves more investigations.

We also observe that a moderate dropout ratio (0.5 for spatial fine-tuning and 0.7 for temporal training) is consistently good. Large dropout ratios like 0.9 may be unsuited for small networks with less layers, particularly under the fine-tuning setting that only adjusts the last several layers. Finally, we find that softmax seems a better choice with consistently good results.

Deep learning based approaches are already showing better results than the traditional techniques for video classification. We believe that the room for further improvement is huge. First, the temporal stream results might be significantly boosted if we could have sufficient labeled video data to train a deeper network. Second, although the two-stream framework is adopted in this work, future approaches do not necessarily need to follow this pipeline. After all, it is all about the way of modeling the temporal dimension of the videos, which can be achieved by using alternative solutions like the RNN or devising new network architectures. 


\section*{Acknowledgments}
{\small The project is supported in part by the National 863 Program (\#2014AA015101), a grant from NSF China (\#61201387), and two grants from the STCSM (\#13PJ1400400, \#13511504503). We thank Jie Shao and Xiaoteng Zhang for their help. } 
\bibliographystyle{abbrv}
\scriptsize
\bibliography{reference}  \end{document}